\documentclass[11pt,a4paper]{article}
\PassOptionsToPackage{breaklinks}{hyperref}
\usepackage[hyperref]{naaclhlt2019}
\usepackage{times}
\usepackage{latexsym}

\usepackage{graphicx}
\usepackage{enumitem}
\usepackage{xcolor}
\usepackage{multirow}
\usepackage{array}
\usepackage{booktabs}
\usepackage{tikz,tikz-qtree}
\usepackage{pgfplots}
\pgfplotsset{compat=1.14}

\definecolor{g-red}{HTML}{DB4437}
\definecolor{g-blue}{HTML}{4285F4}
\definecolor{g-green}{HTML}{0F9D58}
\definecolor{g-yellow}{HTML}{F4B400}
\definecolor{g-orange}{HTML}{FF9800}
\definecolor{g-grey}{HTML}{9E9E9E}

\newcommand{\ignore}[1]{}

\newcolumntype{C}[1]{>{\centering\arraybackslash}p{#1}}

\aclfinalcopy
\title{BoolQ: Exploring the Surprising Difficulty of Natural Yes/No Questions}

\author{Christopher Clark$^{*1}$, Kenton Lee$^\dagger$, Ming-Wei Chang$^\dagger$, Tom Kwiatkowski$^\dagger$\\
\AND
Michael Collins $^{\dagger2}$, Kristina Toutanova$^\dagger$ \\
\\ $^*$Paul G. Allen School of CSE, University of Washington\\
\texttt{\large csquared@cs.uw.edu}\\\\
$^\dagger$Google AI Language \\
\texttt{\large \{kentonl, mingweichang, tomkwiat, mjcollins, kristout\}@google.com} \\
}

\date{December 2018}

\newlength{\savedparindent}
\AtBeginDocument{\setlength{\savedparindent}{\parindent}}

\begin{document}
\maketitle

\begin{abstract}
In this paper we study yes/no questions that are {\em naturally occurring} --- meaning that they are generated in unprompted and unconstrained settings.
We build a reading comprehension dataset, BoolQ, of such questions, and show that they are unexpectedly challenging. They often query for complex, non-factoid information, and require difficult entailment-like inference to solve.
We also explore the effectiveness of a range of transfer learning baselines.
We find that transferring from entailment data is more effective than transferring from paraphrase or extractive QA data, and that it, surprisingly, continues to be very beneficial even when starting from massive pre-trained language models such as BERT.
Our best method trains BERT on MultiNLI and then re-trains it on our train set. It achieves 80.4\% accuracy compared to 90\% accuracy of human annotators (and 62\% majority-baseline), leaving a significant gap for future work.
\end{abstract}

\maketitle

\section{Introduction}
Understanding what facts can be inferred to be true or false from text is an essential part of natural language understanding. In many cases, these inferences can go well beyond what is immediately stated in the text. For example, a simple sentence like ``Hanna Huyskova won the gold medal for Belarus in freestyle skiing." implies that (1) Belarus is a country, (2) Hanna Huyskova is an athlete, (3) Belarus won at least one Olympic event, (4) the USA did \textit{not} win the freestyle skiing event, and so on.
\addtocounter{footnote}{1} 
\footnotetext{Work completed while interning at Google.}
\addtocounter{footnote}{1}
\footnotetext{Also affiliated with Columbia University, work done at Google.}
\begin{figure}[t!]
\begin{small}
\begin{tabular}{p{.35cm}p{6.5cm}}
    \toprule
    \textbf{Q}: & Has the UK been hit by a hurricane? \\
    
    \textbf{P}: & The Great Storm of 1987 was a violent extratropical cyclone which caused casualties in England, France and the Channel Islands \ldots \\
    
    \textbf{A}: & Yes. [An example event is given.] \\
    
    \\
    
    \textbf{Q}: & Does France have a Prime Minister and a President? \\
    
    \textbf{P}: & \ldots~The extent to which those decisions lie with the Prime Minister or President depends upon \ldots \\
    
    \textbf{A}: & Yes. [Both are mentioned, so it can be inferred both exist.] \\
    
    \\
    
    \textbf{Q}: &Have the San Jose Sharks won a Stanley Cup? \\
    \textbf{P}: &\ldots~The Sharks have advanced to the Stanley Cup finals once, losing to the Pittsburgh Penguins in 2016 \ldots \\
    
    \textbf{A}:&  No. [They were in the finals once, and lost.] \\

    \bottomrule    
\end{tabular}
    \end{small}

    \caption{Example yes/no questions from the BoolQ dataset. Each example consists of a question (\textbf{Q}), an excerpt from a passage (\textbf{P}), and an answer (\textbf{A}) with an explanation added for clarity. 
    \label{fig:example}}
\end{figure}

To test a model's ability to make these kinds of inferences, previous work in natural language inference (NLI) proposed the task of labeling candidate statements as being entailed or contradicted by a given passage. However, in practice, generating candidate statements that test for complex inferential abilities is challenging. For instance, evidence suggests~\cite{gururangan2018annotation, jia2017adversarial, mccoy2019right} that simply asking human annotators to write candidate statements will result in examples that typically only require surface-level reasoning.

In this paper we propose an alternative: we test models on their ability to answer naturally occurring yes/no questions.
That is, questions that were authored by people who were not prompted to write particular kinds of questions, including even being required to write yes/no questions, and who did not know the answer to the question they were asking.
Figure~\ref{fig:example} contains some examples from our dataset.
We find such questions often query for non-factoid information, and that human annotators need to apply a wide range of inferential abilities when answering them. As a result, they can be used to construct highly inferential reading comprehension datasets that have the added benefit of being directly related to the practical end-task of answering user yes/no questions.

Yes/No questions do appear as a subset of some existing datasets~\cite{coqa, quac, hotpotqa}. However, these datasets are primarily intended to test other aspects of question answering (QA), such as conversational QA or multi-step reasoning, and do not contain naturally occurring questions.

We follow the data collection method used by Natural Questions (NQ)~\cite{tnq} to gather 16,000 naturally occurring yes/no questions into a dataset we call BoolQ (for Boolean Questions). Each question is paired with a paragraph from Wikipedia that an independent annotator has marked as containing the answer. The task is then to take a question and passage as input, and to return ``yes" or ``no" as output.
Figure~\ref{fig:example} contains some examples, and Appendix~\ref{appendix:random_examples} contains additional randomly selected examples.

Following recent work~\cite{glue}, we focus on using transfer learning to establish baselines for our dataset. Yes/No QA is closely related to many other NLP tasks, including other forms of question answering, entailment, and paraphrasing. Therefore, it is not clear what the best data sources to transfer from are, or if it will be sufficient to just transfer from powerful pre-trained language models such as BERT~\cite{bert} or ELMo~\cite{elmo}.
We experiment with state-of-the-art unsupervised approaches, using existing entailment datasets, three methods of leveraging extractive QA data, and using a few other supervised datasets.

We found that transferring from MultiNLI, and the unsupervised pre-training in BERT, gave us the best results. Notably, we found these approaches are surprisingly complementary and can be combined to achieve a large gain in performance.
Overall, our best model reaches 80.43\% accuracy, compared to 62.31\% for the majority baseline and 90\% human accuracy. In light of the fact BERT on its own has achieved human-like performance on several NLP tasks, this demonstrates the high degree of difficulty of our dataset. We present our data and code at \url{https://goo.gl/boolq}.

\section{Related Work}
Yes/No questions make up a subset of the reading comprehension datasets CoQA~\cite{coqa}, QuAC~\cite{quac}, and HotPotQA~\cite{hotpotqa}, and are present in the ShARC~\cite{sharc} dataset.
These datasets were built to challenge models to understand conversational QA (for CoQA, ShARC and QuAC) or multi-step reasoning (for HotPotQA), which complicates our goal of using yes/no questions to test inferential abilities.
Of the four, QuAC is the only one where the question authors were not allowed to view the text being used to answer their questions, making it the best candidate to contain naturally occurring questions. However, QuAC still heavily prompts users, including limiting their questions to be about pre-selected Wikipedia articles, and is highly class imbalanced with 80\% ``yes" answers.

The MS Marco dataset~\cite{msmarco}, which contains questions with free-form text answers, also includes some yes/no questions. We experiment with  heuristically identifying them in Section~\ref{sect:transfer}, but this process can be noisy and the quality of the resulting annotations is unknown. We also found the resulting dataset is class imbalanced, with 80\% ``yes" answers.

Yes/No QA has been used in other contexts, such as the templated bAbI stories~\cite{babi} or some Visual QA datasets~\cite{vqa, vqa_survey}. We focus on answering yes/no questions using natural language text.

Question answering for reading comprehension in general has seen a great deal of recent work~\cite{squad, triviaqa}, and there have been many recent attempts to construct QA datasets that require advanced reasoning abilities~\cite{hotpotqa, qangaroo, mihaylov2018can, swag, record}. However, these attempts typically involve engineering data to be more difficult by, for example, explicitly prompting users to write multi-step questions~\cite{hotpotqa, mihaylov2018can}, or filtering out easy questions~\cite{swag}. This risks resulting in models that do not have obvious end-use applications since they are optimized to perform in an artificial setting. In this paper, we show that yes/no questions have the benefit of being very challenging even when they are gathered from natural sources.

Natural language inference is also a well studied area of research, particularly on the MultiNLI~\cite{multinli} and SNLI~\cite{snli} datasets. Other sources of entailment data include the PASCAL RTE challenges~\cite{rte6, rte7} or SciTail~\cite{scitail}. We note that, although SciTail, RTE-6 and RTE-7 did not use crowd workers to generate candidate statements, they still use sources (multiple choices questions or document summaries) that were written by humans with knowledge of the premise text.
Using naturally occurring yes/no questions ensures even greater independence between the questions and premise text, and ties our dataset to a clear end-task.
BoolQ also requires detecting entailment in paragraphs instead of sentence pairs.

Transfer learning for entailment has been studied in GLUE~\cite{glue} and SentEval~\cite{senteval}.
Unsupervised pre-training in general has recently shown excellent results on many datasets, including entailment data~\cite{elmo, bert, openai}.

Converting short-answer or multiple choice questions into entailment examples, as we do when experimenting with transfer learning, has been proposed in several prior works~\cite{demszky2018transforming, poliak2018collecting, scitail}. In this paper we found some evidence suggesting that these approaches are less effective than using crowd-sourced entailment examples when it comes to transferring to natural yes/no questions.

Contemporaneously with our work,~\citet{phang2018sentence} showed that pre-training on supervised tasks could be beneficial even when using pre-trained language models, especially for a textual entailment task. Our work confirms these results for yes/no question answering.

This work builds upon the Natural Questions (NQ)~\cite{tnq}, which contains some natural yes/no questions. However, there are too few (about 1\% of the corpus) to make yes/no QA a very important aspect of that task.
In this paper, we gather a large number of additional yes/no questions in order to construct a dedicated yes/no QA dataset.

\section{The BoolQ Dataset}
An example in our dataset consists of a question, a paragraph from a Wikipedia article, the title of the article, and an answer, which is either ``yes" or ``no". We include the article title since it can potentially help resolve ambiguities (e.g., coreferent phrases) in the passage, although none of the models presented in this paper make use of them.

\subsection{Data Collection}
\label{sect:data_collection}

We gather data using the pipeline from NQ~\cite{tnq}, but with an additional filtering step to focus on yes/no questions. We summarize the complete pipeline here, but refer to their paper for a more detailed description.

Questions are gathered from anonymized, aggregated queries to the Google search engine. Queries that are likely to be yes/no questions are heuristically identified: we found selecting queries where the first word is in a manually constructed set of indicator words\footnote{The full set is: \{``did", ``do", ``does", ``is", ``are", ``was",  ``were", ``have", ``has", ``can", ``could", ``will", ``would"\}.}
and are of sufficient length, to be effective.

Questions are only kept if a Wikipedia page is returned as one of the first five results, in which case the question and Wikipedia page are given to a human annotator for further processing.

Annotators label question/article pairs in a three-step process. First, they decide if the question is \textit{good}, meaning it is comprehensible, unambiguous, and requesting factual information. This judgment is made before the annotator sees the Wikipedia page. Next, for good questions, annotators find a passage within the document that contains enough information to answer the question. Annotators can mark questions as ``not answerable" if the Wikipedia article does not contain the requested information. Finally, annotators mark whether the question's answer is ``yes" or ``no". Annotating data in this manner is quite expensive since annotators need to search entire Wikipedia documents for relevant evidence and read the text carefully.

Note that, unlike in NQ, we only use questions that were marked as having a yes/no answer, and pair each question with the selected passage instead of the entire document. This helps reduce ambiguity (ex., avoiding cases where the document supplies conflicting answers in different paragraphs), and keeps the input small enough so that existing entailment models can easily be applied to our dataset.

\begin{table*}[htp!]
\centering
\begin{small}
\begin{tabular}{llrr}

\toprule
       \multicolumn{4}{c}{Question Topic} \\ \toprule
Category            & Example                                             & Percent & Yes\% \\ \hline
Entertainment Media & Is You and I by Lady Gaga a cover?              & 22.0    & 65.9        \\ 
Nature/Science      & Are there blue whales in the Atlantic Ocean?           & 22.0    & 56.8        \\ 
Sports              & Has the US men's team ever won the World Cup? & 11.0     & 54.5        \\       
Law/Government      & Is there a seat belt law in New Hampshire?          & 10.0    & 70.0        \\ 
History             & Were submarines used in the American Civil War?     & 5.0     & 70.0        \\ 
Fictional Events    & Is the Incredible Hulk part of the avengers?      & 4.0     & 87.5        \\ 
Other               & Is GDP per capita same as per capita income?     & 26.0    & 65.4       \\ \toprule  
\multicolumn{4}{c}{Question Type} \\ \toprule
Category            & Example                                             & Percent & Yes\% \\ \hline

Definitional      & Is thread seal tape the same as Teflon tape?       & 14.5    & 55.2        \\ 
Existence         & Is there any dollar bill higher than a 100? & 14.5    & 69.0 \\
Event Occurrence  & Did the great fire of London destroy St. Paul's Cathedral?      & 11.5    & 73.9        \\
Other General Fact      & Is there such thing as a dominant eye?             & 29.5    & 62.7        \\ 
Other Entity Fact       & Is the Arch in St. Louis a national park?          & 30.0    & 63.3        \\ 
\bottomrule
\end{tabular}
\end{small}

\caption{Question categorization of BoolQ. Question topics are shown in the top half and question types are shown in the bottom half.\label{fig:question_categories}}
\end{table*}

\ignore{
Unlike in NQ, we only make use of questions that were marked as having a yes/no answer and exclude questions if the annotator selected a list or table as the passage. We pair each question with the passage selected by the annotators instead of the entire document to help reduce ambiguity (ex., avoiding cases where the document supplies conflicting answers in different paragraphs), and to keep the input small enough so that existing entailment models can easily be applied to our dataset. We leave the full-document setting to future work.
}

We combine 13k questions gathered from this pipeline with an additional 3k questions with yes/no answers from the NQ training set to reach a total of 16k questions. We split these questions into a 3.2k dev set, 3.2k test set, and 9.4k train set, ensuring questions from NQ are always in the train set.
``Yes'' answers are slightly more common (62.31\% in the train set). The queries are typically short (average length 8.9 tokens) with longer passages (average length 108 tokens).

\subsection{Analysis}

In the following section we analyze our dataset to better understand the nature of the questions, the annotation quality, and the kinds of reasoning abilities required to answer them.

\subsection{Annotation Quality}
First, in order to assess annotation quality, three of the authors labelled 110 randomly chosen examples. If there was a disagreement, the authors conferred and selected a single answer by mutual agreement.
We call the resulting labels ``gold-standard" labels.
On the 110 selected examples, the answer annotations reached 90\% accuracy compared to the gold-standard labels.
Of the cases where the answer annotation differed from the gold-standard, six were ambiguous or debatable cases, and five were errors where the annotator misunderstood the passage.
Since the agreement was sufficiently high, we elected to use singly-annotated examples in the training/dev/test sets in order to be able to gather a larger dataset.
\newcommand{\tmini}[1]{\multirow{2}{*}{
        \begin{minipage}{5cm}#1\end{minipage}}}
\begin{table*}[ht]
\begin{small}
    \centering
    \begin{tabular}{p{5cm}lp{9.5cm}}
    \toprule
        Reasoning Types & \multicolumn{2}{c}{Yes/No Question Answering Examples} \\ \hline
        
        \textbf{Paraphrasing} (38.7\%) 
            & {\bf Q}:&  Is Tim Brown in the Hall of Fame? \\  
        \tmini{The passage explicitly asserts or refutes what is stated in the question.} 
            & {\bf P}:& Brown has also played for the Tampa Bay Buccaneers. In 2015, he was inducted into the Pro Football Hall of Fame. \\
            & {\bf A}:& Yes. [``inducted into" directly implies he is in Hall of Fame.]
    \\ \hline
        \textbf{By Example} (11.8\%)
            & {\bf Q}: & Are there any nuclear power plants in Michigan? \\  
        \tmini{The passage provides an example or counter-example to what is asserted by the question.} 
            & {\bf P}: & \ldots three nuclear power plants supply Michigan with about 30\% of its electricity.\\
            & {\bf A}: &Yes. [Since there must be at least three.]
    \\ \hline
        \textbf{Factual Reasoning} (8.5\%) 
            & {\bf Q}: & Was designated survivor filmed in the White House?  \\  
        \tmini{Answering the question requires using world-knowledge to connect what is stated in the passage to the question. }
            & {\bf P}: & The series is\ldots filmed in Toronto, Ontario. \\
            & {\bf A}: & No. [The White House is not located in Toronto.] \\
    \\ \hline
        \textbf{Implicit} (8.5\%) 
            & {\bf Q}: & Is static pressure the same as atmospheric pressure? \\  
        \tmini{The passage mentions or describes entities in the question in way that would not make sense if the answer was not yes/no.} 
            & {\bf P}: & The aircraft designer's objective is to ensure the pressure in the aircraft's static pressure system is as close as possible to the atmospheric pressure\ldots \\
            & {\bf A}: & No. [It would not make sense to bring them ``as close as possible" if those terms referred to the same thing.] 
    \\ \hline
        \textbf{Missing Mention} (6.6\%)
            & {\bf Q}: & Did Bonnie Blair's daughter make the Olympic team? \\ 
        \tmini{We can conclude the answer is yes or no because, if this was not the case, it would have been mentioned in the passage.}
            & {\bf P}: & Blair and Cruikshank have two children: a son, Grant, and daughter, Blair.... Blair Cruikshank competed at the 2018 United States Olympic speed skating trials at the 500 meter distance.\\
            & {\bf A}: & No. [The passage describes Blair Cruikshank's daughter's skating accomplishments, so it would have mentioned it if she had qualified.] 
    \\ \hline
        \textbf{Other Inference} (25.9\%)
            & {\bf Q}: & Is the sea snake the most venomous snake? \\
        \tmini{The passage states a fact that can be used to infer whether the answer is true or false, and does not fall into any of the other categories.}
            & {\bf P}: &\ldots the venom of the inland taipan, drop by drop, is the most toxic among all snakes\\
            & {\bf A}: & No. [If inland taipan is the most venomous snake, the sea snake must not be.]             
            \\
        \bottomrule
    \end{tabular}
    \end{small}
    \caption{Kinds of reasoning needed in the BoolQ dataset.\label{tab:reasoning}}
\end{table*}

\subsection{Question Types}
Part of the value of this dataset is that it contains questions that people genuinely want to answer. To explore this further, we manually define a set of topics that questions can be about. An author categorized 200 questions into these topics. The results can be found in the upper half of Table~\ref{fig:question_categories}.

Questions were often about entertainment media (including T.V., movies, and music), along with other popular topics like sports. However, there are still a good portion of questions asking for more general factual knowledge, including ones about historical events or the natural world.

We also broke the questions into categories based on what kind of information they were requesting, shown in the lower half of Table~\ref{fig:question_categories}. Roughly one-sixth of the questions are about whether anything with a particular property exists (Existence), another sixth are about whether a particular event occurred (Event Occurrence), and another sixth ask whether an object is known by a particular name, or belongs to a particular category (Definitional). The questions that do not fall into these three categories were split between requesting facts about a specific entity, or requesting more general factual information.

We do find a correlation between the nature of the question and the likelihood of a ``yes" answer. However, this correlation is too weak to help outperform the majority baseline because, even if the topic or type is known, it is never best to guess the minority class. We also found that question-only models perform very poorly on this task (see Section \ref{sect:one-sided}), which helps confirm that the questions do not contain sufficient information to predict the answer on their own.

\subsection{Types of Inference}
\label{sect:inferences}

Finally, we categorize the kinds of inference required to answer the questions in BoolQ\footnote{Note the dataset has been updated since we carried out this analysis, so it might be slighly out-of-date.}. The definitions and results are shown in Table~\ref{tab:reasoning}.

Less than 40\% of the examples can be solved by detecting paraphrases. Instead, many questions require making additional inferences (categories ``Factual Reasoning", ``By Example", and ``Other Inference") to connect what is stated in the passage to the question. There is also a significant class of questions (categories ``Implicit" and ``Missing Mention") that require a subtler kind of inference based on how the passage is written.

\subsection{Discussion}
Why do natural yes/no questions require inference so often? We hypothesize that there are several factors. First, we notice factoid questions that ask about simple properties of entities, such as ``Was Obama born in 1962?", are rare. We suspect this is because people will almost always prefer to phrase such questions as short-answer questions (e.g., ``When was Obama born?"). Thus, there is a natural filtering effect where people tend to use yes/no questions exactly when they want more complex kinds of information.

Second, both the passages and questions rarely include negation. As a result, detecting a ``no" answer typically requires understanding that a positive assertion in the text excludes, or makes unlikely, a positive assertion in the question. This requires reasoning that goes beyond paraphrasing (see the ``Other-Inference" or ``Implicit" examples).

We also think it was important that annotators only had to answer questions, rather than generate them. For example, imagine trying to construct questions that fall into the categories of ``Missing Mention" or ``Implicit". While possible, it would require a great deal of thought and creativity. On the other hand, detecting when a yes/no question can be answered using these strategies seems much easier and more intuitive. Thus, having annotators answer pre-existing questions opens the door to building datasets that contain more inference and have higher quality labels.

\section{Training Yes/No QA Models}
\label{sect:transfer}
Models on this dataset need to predict an output class given two pieces of input text, which is a well studied paradigm~\cite{glue}.
We find training models on our train set alone to be relatively ineffective. Our best model reaches 69.6\% accuracy, only 8\% better than the majority baseline. Therefore, we follow the recent trend in NLP of using transfer learning. In particular, we experiment with {\em pre-training} models on related tasks that have larger datasets, and then {\em fine-tuning} them on our training data. We list the sources we consider for pre-training below.
\newline
\newline
\noindent
\textbf{Entailment:} We consider two entailment  datasets, \textit{MultiNLI}~\cite{multinli} and ~\textit{SNLI}~\cite{snli}. We choose these datasets since they are widely-used and large enough to use for pre-training. We also experiment with ablating classes from MultiNLI. During fine-tuning we use the probability the model assigns to the ``entailment" class as the probability of predicting a ``yes" answer.
\newline
\newline
\noindent
\textbf{Multiple-Choice QA:}
We use a multiple choice reading comprehension dataset, \textit{RACE}~\cite{race}, which contains stories or short essays paired with questions built to test the reader's comprehension of the text.
Following what was done in SciTail~\cite{scitail}, we convert questions and answer-options to statements by either substituting the answer-option for the blanks in fill-in-the-blank questions, or appending a separator token and the answer-option to the question.
During training, we have models independently assign a score to each statement, and then apply the softmax operator between all statements per each question to get statement probabilities. We use the negative log probability of the correct statement as a loss function. To fine-tune on BoolQ, we apply the sigmoid operator to the score of the question given its passage to get the probability of a ``yes" answer.
\newline
\newline
\noindent
\textbf{Extractive QA:}
We consider several methods of leveraging extractive QA datasets, where the model must answer questions by selecting text from a relevant passage. Preliminary experiments found that simply transferring the lower-level weights of extractive QA models was ineffective, so we instead consider three methods of constructing entailment-like data from extractive QA data.

First, we use the \textit{QNLI} task from GLUE~\cite{glue}, where the model must determine if a sentence from SQuAD 1.1~\cite{squad} contains the answer to an input question or not.
Following previous work~\cite{hu2018read_verify}, we also try building entailment-like training data from \textit{SQuAD 2.0}~\cite{squad2}. We concatenate questions with either the correct answer, or with the incorrect ``distractor" answer candidate provided by the dataset, and train the model to classify which is which given the question's supporting text.

Finally, we also experiment with leveraging the long-answer portion of NQ, where models must select a paragraph containing the answer to a question from a document. Following our method for Multiple-Choice QA, we train a model to assign a score to (question, paragraph) pairs, apply the softmax operator on paragraphs from the same document to get a probability distribution over the paragraphs, and train the model on the negative log probability of selecting an answer-containing paragraph. We only train on questions that were marked as having an answer, and select an answer-containing paragraph and up to 15 randomly chosen non-answer-containing paragraphs for each question. On BoolQ, we compute the probability of a ``yes" answer by applying the sigmoid operator to the score the model gives to the input question and passage.
\newline
\newline
\noindent
\textbf{Paraphrasing:} We use the Quora Question Paraphrasing (\textit{QQP}) dataset, which consists of pairs of questions labelled as being paraphrases or not.\footnote{data.quora.com/First-Quora-Dataset-Release-Question-Pairs} Paraphrasing is related to entailment since we expect, at least in some cases, passages will contain a paraphrase of the question.
\newline
\newline
\noindent
\textbf{Heuristic Yes/No:} We attempt to heuristically construct a corpus of yes/no questions from the MS Marco corpus~\cite{msmarco}. MS Marco has free-form answers paired with snippets of related web documents. We search for answers starting with ``yes" or ``no", and then pair the corresponding questions with snippets marked as being related to the question. We call this task \textit{Y/N MS Marco}; in total we gather 38k examples, 80\% of which are ``yes'' answers.
\newline
\newline
\noindent
\textbf{Unsupervised:} It is well known that unsupervised pre-training using language-modeling objectives~\cite{elmo, bert, openai}, can improve performance on many tasks. We experiment with these methods by using the pre-trained models from \textit{ELMo}, \textit{BERT}, and OpenAI's Generative Pre-trained Transformer (\textit{OpenAI GPT}) (see Section~\ref{sect:neural_models}).

\section{Results}

\subsection{Shallow Models}
First, we experiment with using a linear classifier on our task. In general, we found features such as word overlap or TF-IDF statistics were not sufficient to achieve better than the majority-class baseline accuracy (62.17\% on the dev set). We did find there was a correlation between the number of times question words occurred in the passage and the answer being ``yes", but the correlation was not strong enough to build an effective classifier. ``Yes" is the most common answer even among questions with zero shared words between the question and passage (with a 51\% majority), and more common in other cases.

\subsection{Neural Models}
\label{sect:neural_models}
For our experiments that do not use unsupervised pre-training (except the use of pre-trained word vectors), we use a standard recurrent model with attention. Our experiments using unsupervised pre-training use the models provided by the authors. In more detail:

Our \textit{Recurrent} model follows a standard recurrent plus attention architecture for text-pair classification~\cite{glue}. It embeds the premise/hypothesis text using fasttext word vectors~\cite{fasttext_word_vectors} and learned character vectors, applies a shared bidirectional LSTM to both parts, applies co-attention~\cite{parikh2016decomposable} to share information between the two parts, applies another bi-LSTM to both parts, pools the result, and uses the pooled representation to predict the final class. See Appendix~\ref{app:model} for details.

Our \textit{Recurrent \begin{small}+\end{small}ELMo} model uses the language model from~\citet{elmo} to provide contextualized embeddings to the baseline model outlined above, as recommended by the authors.

Our \textit{OpenAI GPT} model fine-tunes the 12 layer 768 dimensional uni-directional transformer from~\citet{openai}, which has been pre-trained as a language model on the Books corpus~\cite{books}.

Our \textit{BERT\textsubscript{L}} model fine-tunes the 24 layer 1024 dimensional transformer from~\citet{bert}, which has been trained on next-sentence-selection and masked language modelling on the Book Corpus and Wikipedia.

We fine-tune the BERT\textsubscript{L} and the OpenAI GPT models using the optimizers recommended by the authors, but found it important to tune the optimization parameters to achieve the best results. We use a batch size of 24, learning rate of 1e-5, and 5 training epochs for BERT and a learning rate of 6.25e-5, batch size of 6, language model loss of 0.5, and 3 training epochs for OpenAI GPT.

\subsection{Question/Passage Only Results}
\label{sect:one-sided}
Following the recommendation of~\citet{gururangan2018annotation}, we first experiment with models that are only allowed to observe the question or the passage. The pre-trained BERT\textsubscript{L} model reached 64.48\% dev set accuracy using just the question and 66.74\% using just the passage.
Given that the majority baseline is 62.17\%, this suggests there is little signal in the question by itself, but that some language patterns in the passage correlate with the answer.
Possibly, passages that present more straightforward factual information (like Wikipedia introduction paragraphs) correlate with ``yes" answers.

\subsection{Transfer Learning Results}

\begin{table*}
    \centering
    \begin{tabular}{lllcC{10mm}C{9mm}}
        \toprule
         Transfer Task & Model & Transfer Data & \#Examples & Source Acc. & BoolQ Acc. \\ 
         \toprule
         N/A & Majority & - & - & - & 62.17  \\  
         N/A  & Recurrent & - & - & - &  69.60\\ \hline \hline
         \multirow{3}{*}{Extractive QA} & \multirow{3}{*}{Recurrent} & QNLI & 108k & 79.66 &  71.36\\ 
         & & SQuAD 2.0 & 130k & 69.45 & 69.83 \\
         & & NQ Long Answer & 93k & 71.78 & 72.78 \\ \hline
         Paraphrasing & Recurrent & QQP & 364k & 89.58 & 71.30 \\\hline
         Heuristic Y/N & Recurrent &Y/N MS Marco & 39k & 87.26 &  71.40 \\ \hline
         \multirow{4}{*}{Entailment} & \multirow{4}{*}{Recurrent} & MultiNLI & 392k &  78.23 & 75.57 \\
         & & \hspace{3mm}- w/o Entail & 262k & 84.26 & 72.95\\
         & & \hspace{3mm}- w/o Contradict & 262k & 81.16 &  72.85 \\
         & & \hspace{3mm}- w/o Neutral & 262k & 89.72 & 74.83 \\
         & & SNLI & 351k & 88.17 & 73.16 \\ \hline
         MC QA & Recurrent & RACE & 549k & 42.30 & 68.40 \\  \hline
         \multirow{3}{*}{Unsupervised} & Recurrent \begin{small}+\end{small}ELMo & Billion Word & 1000M & - & 71.41 \\ 
          & OpenAI GPT  & Books & 800M & - & 72.87 \\ 
         & BERT\textsubscript{L} & Books/Wikipedia & 3,300M & - & 76.90 \\
         \bottomrule
    \end{tabular}
    \caption{Transfer learning results on the BoolQ dev set after fine-tuning on
    the BoolQ training set. Results are averaged over five runs. In all cases directly using the pre-trained model without fine-tuning did not achieve results better than the majority baseline, so we do not include them here.\label{tab:transfer_results}}
\end{table*}

The results of our transfer learning methods are shown in Table~\ref{tab:transfer_results}. All results are averaged over five runs. For models pre-trained on supervised datasets, both the pre-training and the fine-tuning stages were repeated. For unsupervised pre-training, we use the pre-trained models provided by the authors, but continue to average over five runs of fine-tuning.
\newline
\newline
\noindent
\textbf{QA Results:}
We were unable to transfer from RACE or SQuAD 2.0. For RACE, the problem might be domain mismatch. In RACE the passages are stories, and the questions often query for passage-specific information such as the author's intent or the state of a particular entity from the passage, instead of general knowledge.

We would expect SQuAD 2.0 to be a better match for BoolQ since it is also Wikipedia-based, but its possible detecting the adversarially-constructed distractors used for negative examples does not relate well to yes/no QA.

We got better results using QNLI, and even better results using NQ. This shows the task of selecting text relevant to a question is partially transferable to yes/no QA, although we are only able to gain a few points over the baseline.
\newline
\newline
\noindent
\textbf{Entailment Results:}
The MultiNLI dataset out-performed all other supervised methods by a large margin. Remarkably, this approach is only a few points behind BERT despite using orders of magnitude less training data and a much more light-weight model, showing high-quality pre-training data can help compensate for these deficiencies.

Our ablation results show that removing the neutral class from MultiNLI hurt transfer slightly, and removing either of the other classes was very harmful, suggesting the neutral examples had limited  value.
SNLI transferred better than other datasets, but worse than MultiNLI. We suspect this is due to limitations of the photo-caption domain it was constructed from.
\newline
\newline
\noindent
\textbf{Other Supervised Results:}
We obtained a small amount of transfer using QQP and Y/N MS Marco. Although Y/N MS Marco is a yes/no QA dataset, its small size and class imbalance likely contributed to its limited effectiveness. The web snippets it uses as passages also present a large domain shift from the Wikipedia passages in BoolQ.
\newline
\newline
\noindent
\textbf{Unsupervised Results:}
Following results on other datasets~\cite{glue}, we found BERT\textsubscript{L} to be the most effective unsupervised method, surpassing all other methods of pre-training.
\subsection{Multi-Step Transfer Results}

\begin{table}
    \centering
    \begin{tabular}{l|c|c}
    \toprule
         Model & Dev Acc. & Test Acc. \\ \toprule
         Majority Class &62.17 & 62.31\\
         Recurrent & 70.28 & 67.52\\
         \hspace{2mm}\begin{small}+\end{small}MultiNLI & 76.15& 74.24\\
         Pre-trained BERT\textsubscript{L} & 78.09 & 76.70 \\
         \hspace{2mm}\begin{small}+\end{small}MultiNLI & 82.20 & 80.43\\
    \bottomrule
    \end{tabular}
    \caption{Test set results on BoolQ, ``+MultiNLI" indicates models that were additionally pre-trained on MultiNLI before being fine-tuned on the train set.\label{tab:test_results}}
\end{table}

Our best single-step transfer learning results were from using the pre-trained BERT\textsubscript{L} model and MultiNLI. We also experiment with combining these approaches using a two-step pre-training regime. In particular, we fine-tune the pre-trained BERT\textsubscript{L} on MultiNLI, and then fine-tune the resulting model again on the BoolQ train set. We found decreasing the number of training epochs to 3 resulted in a slight improvement when using the model pre-trained on MultiNLI.

We show the test set results for this model, and some other pre-training variations, in Table~\ref{tab:test_results}. For these results we train five versions of each model using different training seeds, and show the model that had the best dev-set performance.

Given how extensively the BERT\textsubscript{L} model has been pre-trained, and how successful it has been across many NLP tasks, the additional gain of 3.5 points due to using MultiNLI is remarkable. This suggests MultiNLI contains signal orthogonal to what is found in BERT's unsupervised objectives.

\subsection{Sample Efficiency}

\begin{figure}[b]
\begin{tikzpicture}
 \begin{axis}[
   width=1.0\columnwidth,
   height=0.75\columnwidth,
   legend columns=2,
   legend style={at={(0.0, -0.4)},anchor=south west,font=\scriptsize},
   font=\scriptsize,
   xmin=970, xmax=9426,
   ymin=60, ymax=84,
  xtick={1000,2000,3000,4000,5000,6000,7000,8000,9000,10000},
   ymajorgrids=true,
   xmajorgrids=true,
   xlabel style={yshift=0.5ex,},
   xlabel=Number of Training Examples,
   ylabel=BoolQ Dev Accuracy,
   ylabel style={yshift=-0.5ex,}]

\addplot[mark=*,g-blue,mark options={scale=1.0}] plot coordinates {
(1000, 61.4801228046)
(2000, 62.3792052269)
(4000, 65.4311919212)
(6000, 67.2599387169)
(8000, 68.4587156773)
(9426, 69.602445364)
};
\addlegendentry{Recurrent}

\addplot[mark=triangle,g-red] plot coordinates {
(1000, 63.1434) 
(2000, 66.4828) 
(4000, 71.2929) 
(6000, 75.3983) 
(8000, 75.4289)
(9426, 76.90566)
};
\addlegendentry{BERT\textsubscript{L}}

\addplot[g-blue,dashed,mark=*] plot coordinates {
(1000, 68.366972208)
(2000, 70.0856268406)
(4000, 72.0428133011)
(6000, 73.9510703087)
(8000, 75.3333330154)
(9426, 75.5657494068
)
};
\addlegendentry{Recurrent +MultiNLI}

\addplot[mark=triangle,g-red,dashed] plot coordinates {
(1000, 73.8971) 
(2000, 76.5012) 
(4000, 78.4007) 
(6000, 79.9326) 
(8000, 80.3615)
(9426, 81.76472)
};
\addlegendentry{BERT\textsubscript{L} +MultiNLI}

     \end{axis}
\end{tikzpicture}
\caption{Accuracy for various models on the BoolQ dev set as the number of training examples varies.}

\label{fig:step_abalation}
\end{figure}
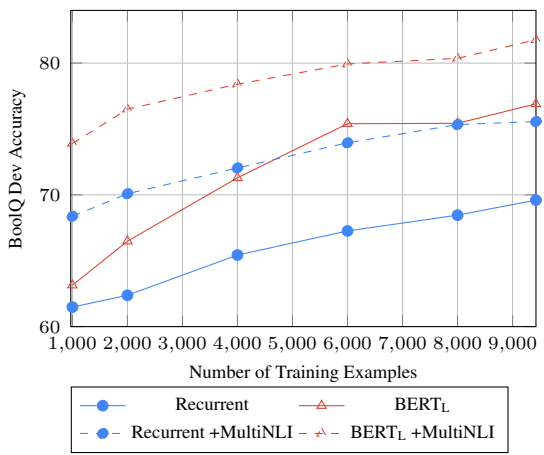

\ignore{
\begin{figure}
\begin{tikzpicture}
 \begin{axis}[
   width=0.95\columnwidth,
   height=0.7\columnwidth,
   legend cell align=left,
   legend style={at={(1, 0)},anchor=south east,font=\scriptsize},
   mark options={mark size=3},
   font=\scriptsize,
   xmin=0, xmax=1000,
   ymin=75, ymax=85,
   xtick={200,400,600,800,1000},
   ymajorgrids=true,
   xmajorgrids=true,
   xlabel style={yshift=0.5ex,},
   xlabel=Pre-training Steps (Thousands) ,
   ylabel=MNLI Dev Accuracy,
   ylabel style={yshift=-0.5ex,}]
    \addplot[mark=triangle,g-blue] plot coordinates {
      (30, 78.6)
      (50, 79.6)
      (100, 80.5)
      (200, 82.2)
      (400, 83.2)
      (600, 84.0)
      (800, 84.3)
      (1000, 84.4)
    };
    \addlegendentry{(Masked LM)}
    \addplot[mark=x,g-red] plot coordinates {
      (30, 79.4)
      (50, 79.8)
      (100, 80.3)
      (200, 81.0)
      (400, 81.7)
      (600, 81.9)
      (800, 82.1)
      (1000, 82.2)
    };
    \addlegendentry{(Left-to-Right)}
     \end{axis}
\end{tikzpicture}
\caption{Ablation over number of training steps. This shows the MNLI accuracy after fine-tuning, starting from model parameters that have been pre-trained for $k$ steps. The x-axis is the value of $k$.
\label{fig:learning-curve}
}
\end{figure}
}

In Figure 2, we graph model accuracy as more of the training data is used for fine-tuning, both with and without initially pre-training on MultiNLI. Pre-training on MultiNLI gives at least a 5-6 point gain, and nearly a 10 point gain for BERT\textsubscript{L} when only using 1000 examples. For small numbers of examples, the recurrent model with MultiNLI pre-training actually out-performs BERT\textsubscript{L}.

\subsection{Discussion}
A surprising result from our work is that the datasets that more closely resemble the format of BoolQ, meaning they contain questions and multi-sentence passages, such as SQuAD 2.0, RACE, or Y/N MS Marco, were not very useful for transfer. The entailment datasets were stronger despite consisting of sentence pairs. This suggests that adapting from sentence-pair input to question/passage input was not a large obstacle to achieving transfer. Preliminary work found attempting to convert the yes/no questions in BoolQ into declarative statements did not improve transfer from MultiNLI, which supports this hypothesis.

The success of MultiNLI might also be surprising given recent concerns about the generalization abilities of models trained on it~\cite{glockner2018breaking}, particularly related to ``annotation artifacts" caused by using crowd workers to write the hypothesis statements~\cite{gururangan2018annotation}. We have shown that, despite these weaknesses, it can still be an important starting point for models being used on natural data.

We hypothesize that a key advantage of MultiNLI is that it contains examples of contradictions. The other sources of transfer we consider, including the next-sentence-selection objective in BERT, are closer to providing examples of entailed text vs. neutral/unrelated text. Indeed, we found that our two step transfer procedure only reaches 78.43\% dev set accuracy if we remove the contradiction class from MultiNLI, regressing its performance close to the level of BERT\textsubscript{L} when just using unsupervised pre-training.

Note that it is possible to pre-train a model on several of the suggested datasets, either in succession or in a multi-task setup. We leave these experiments to future work.
Our results also suggest pre-training on MultiNLI would be helpful for other corpora that contain yes/no questions.

\section{Conclusion}
We have introduced BoolQ, a new reading comprehension dataset of naturally occurring yes/no questions. We have shown these questions are challenging and require a wide range of inference abilities to solve.
We have also studied how transfer learning performs on this task, and found crowd-sourced entailment datasets can be leveraged to boost performance even on top of language model pre-training.
Future work could include building a document-level version of this task, which would increase its difficulty and its correspondence to an end-user application.

\bibliographystyle{acl_natbib}
\bibliography{bibliography.bib}

\pagenumbering{arabic}

\appendix
\section{Appendices}
\subsection{Randomly Selected Examples}
\label{appendix:random_examples}
We include a number of randomly selected examples from the BoolQ train set in Figure~\ref{fig:random_examples}. For each example we show the question in bold, followed by the answer in parentheses, and then the passage below.

\subsection{Recurrent Model}
\label{app:model}
Our recurrent model is a standard model from the text pair classification literature, similar to the one used in the GLUE baseline~\cite{glue} and the model from~\citet{chen2016enhanced}. Our model has the following stages:
\\
\\
\noindent
\textbf{Embed}: Embed the words using a character CNN following what was done by \citet{bidaf}, and the fasttext crawl word embeddings~\cite{fasttext_word_vectors}. Then run a BiLSTM over the results to get context-aware word hypothesis embeddings $\langle u_1, u_2, u_3, ... \rangle$ and premise embeddings $\langle v_1, v_2, v_3, ... \rangle$.
\\
\\
\noindent
\textbf{Co-Attention}: Compute a co-attention matrix, $A$, between the hypothesis and premise where $A_{ij} = w_1 \cdot u_i + w_2 \cdot v_j + w_3 \cdot (u_i \circ v_j)$, $\circ$ is elementwise multiplication, and $w_1$, $w_2$, and $w_3$ are weights to be learned.
\\
\\
\noindent
\textbf{Attend}: For each row in $A$, apply the softmax operator and use the results to compute a weighed sum of the hypothesis embeddings, resulting in attended vectors $\langle \tilde{u}_1, \tilde{u}_2, ... \rangle$. We use the transpose of $A$ to compute vectors $\langle \tilde{v}_1, \tilde{v}_2, ... \rangle$ from the premise embeddings in a similar manner.
\\
\\
\noindent
\textbf{Pool}: Run another BiLSTM over $\langle [v_1; \tilde{v}_1; \tilde{v}_1 \circ v_1], [v_2; \tilde{v}_2; \tilde{v}_2 \circ v_2], ... \rangle$ to get embeddings $\langle h_1, h_2, ...\rangle$. Then pool these embeddings by computing attention scores $a_i =w \cdot h_i$, $p = softmax(a)$, and then the sum $v^*$ = $\sum_i p_ih_i$. Likewise we compute $p^*$ from the premise.
\\
\\
\noindent
\textbf{Classify}: Finally we feed $[v^*; p^*]$ into a fully connected layer, and then through a softmax layer to predict the output class.
\\
\\
We apply dropout at a rate of 0.2 between all layers, and train the model using the Adam optimizer~\cite{adam}. The learning rate is decayed by 0.999 every 100 steps. We use 200 dimensional LSTMs and a 100 dimensional fully connected layer.

\begin{figure*}[h]

\begin{small}

\textbf{Is there a catalytic converter on a diesel? (Y)}
\newline
\hspace*{\savedparindent}A catalytic converter is an exhaust emission control device that converts toxic gases and pollutants in exhaust gas from an internal combustion engine into less-toxic pollutants by catalyzing a redox reaction (an oxidation and a reduction reaction). Catalytic converters are usually used with internal combustion engines fueled by either gasoline or diesel--including lean-burn engines as well as kerosene heaters and stoves.
\\ \\
\textbf{Is there a season 2 of Pride and Prejudice? (N)}
\newline
\hspace*{\savedparindent}Pride and Prejudice is a six-episode 1995 British television drama, adapted by Andrew Davies from Jane Austen's 1813 novel of the same name. Jennifer Ehle and Colin Firth starred as Elizabeth Bennet and Mr. Darcy. Produced by Sue Birtwistle and directed by Simon Langton, the serial was a BBC production with additional funding from the American A\&E Network. BBC1 originally broadcast the 55-minute episodes from 24 September to 29 October 1995. The A\&E Network aired the series in double episodes on three consecutive nights beginning 14 January 1996. There are six episodes in the series.
\\ \\
\textbf{Is Saving Private Ryan based on a book? (N)}
\newline
\hspace*{\savedparindent}In 1994, Robert Rodat wrote the script for the film. Rodat's script was submitted to producer Mark Gordon, who liked it and in turn passed it along to Spielberg to direct. The film is loosely based on the World War II life stories of the Niland brothers. A shooting date was set for June 27, 1997.
\\ \\
\textbf{Is The Talk the same as The View? (N)}
\newline
\hspace*{\savedparindent}In November 2008, the show's post-election day telecast garnered the biggest audience in the show's history at 6.2 million in total viewers, becoming the week's most-watched program in daytime television. It was surpassed on July 29, 2010, during which former President Barack Obama first appeared as a guest on The View, which garnered a total of 6.6 million viewers. In 2013, the show was reported to be averaging 3.1 million daily viewers, which outpaced rival talk show The Talk.
\\ \\
\textbf{Does the concept of a contact force apply to both a macroscopic scale and an atomic scale? (N)}
\newline
\hspace*{\savedparindent}In the Standard Model of modern physics, the four fundamental forces of nature are known to be non-contact forces. The strong and weak interaction primarily deal with forces within atoms, while gravitational effects are only obvious on an ultra-macroscopic scale. Molecular and quantum physics show that the electromagnetic force is the fundamental interaction responsible for contact forces. The interaction between macroscopic objects can be roughly described as resulting from the electromagnetic interactions between protons and electrons of the atomic constituents of these objects. Everyday objects do not actually touch; rather, contact forces are the result of the interactions of the electrons at or near the surfaces of the objects.
\\ \\
\textbf{Legal to break out of prison in Germany? (Y)}
\newline
\hspace*{\savedparindent}In Mexico, Belgium, Germany and Austria, the philosophy of the law holds that it is human nature to want to escape. In those countries, escapees who do not break any other laws are not charged for anything and no extra time is added to their sentence. However, in Mexico, officers are allowed to shoot prisoners attempting to escape, and an escape is illegal if violence is used against prison personnel or property, or if prison inmates or officials aid the escape.
\\ \\
\textbf{Is the movie sand pebbles based on a true story? (N)}
\newline
\hspace*{\savedparindent}The Sand Pebbles is a 1966 American war film directed by Robert Wise in Panavision. It tells the story of an independent, rebellious U.S. Navy machinist's mate, first class aboard the fictional gunboat USS San Pablo in 1920s China.
\\ \\
\textbf{Is Burberrys of London the same as Burberry? (Y)}
\newline
\hspace*{\savedparindent}Burberry was founded in 1856 when 21-year-old Thomas Burberry, a former draper's apprentice, opened his own store in Basingstoke, Hampshire, England. By 1870, the business had established itself by focusing on the development of outdoors attire. In 1879, Burberry introduced in his brand the gabardine, a hardwearing, water-resistant yet breathable fabric, in which the yarn is waterproofed before weaving. ``Burberry'' was the original name until it became ``Burberrys'', due to many customers from around the world began calling it ``Burberrys of London''. In 1999, the name was reverted to the original, ``Burberry''. However, the name ``Burberrys of London'' is still visible on many older Burberry products. In 1891, Burberry opened a shop in the Haymarket, London. Before being termed as trench, it was known as the Tielocken worn by the British officers and featured a belt with no buttons, was double breasted, and protected the body from neck to knees.
\\ \\
\textbf{Is the Saturn Vue the same as the Chevy Equinox? (N)}
\newline
\hspace*{\savedparindent}Riding on the GM Theta platform, the unibody is mechanically similar to the Saturn Vue and the Suzuki XL7. However, the Equinox and the Torrent are larger than the Vue, riding on a 112.5 in (2,858mm) wheelbase, 5.9 in (150mm) longer than the Vue. Front-wheel drive is standard, with optional all-wheel drive. They are not designed for serious off-roading like the truck-based Chevrolet Tahoe and Chevrolet TrailBlazer.
\\ \\
\textbf{Is Destin FL on the Gulf of Mexico? (Y)}
\newline
\hspace*{\savedparindent}The city is located on a peninsula separating the Gulf of Mexico from Choctawhatchee Bay. The peninsula was originally an island; hurricanes and sea level changes gradually connected the island to the mainland.

\end{small}
    \caption{Randomly sampled examples from the BoolQ train set.}
    \label{fig:random_examples}
\end{figure*}

\end{document}